\documentclass[10pt,journal,compsoc]{IEEEtran}

\usepackage{microtype}
\usepackage{graphicx}
\usepackage{subfigure}
\usepackage{booktabs} 
\usepackage[pagebackref=false,breaklinks=true,letterpaper=true,bookmarks=false,citecolor={black}, colorlinks=true,linkcolor={black}]{hyperref}
\usepackage{epsfig}
\usepackage{graphicx}
\usepackage{amsmath}
\usepackage{amssymb}
\usepackage{color}
\usepackage[normalem]{ulem}
\usepackage{calligra}
\usepackage{multirow}
\usepackage{verbatim}
\usepackage[linesnumbered,ruled]{algorithm2e}
\usepackage{bbm}
\usepackage{amsmath} 
\usepackage{amssymb}  

\definecolor{mayablue}{rgb}{0.45, 0.76, 0.98}
\definecolor{burntorange}{rgb}{0.8, 0.33, 0.0}

\newcommand{\eg}{\textit{e.g.}}
\newcommand{\ie}{\textit{i.e.}}

\newcommand{\etal}{\textit{et al.}}

\ifCLASSOPTIONcompsoc
  \usepackage[nocompress]{cite}
\else
  \usepackage{cite}
\fi

\ifCLASSINFOpdf
\else

\fi

\begin{document}
\title{Similarity-preserving Image-image Domain Adaptation for Person Re-identification} 


\author{Weijian~Deng,
        Liang Zheng,
        Qixiang~Ye,
        Yi Yang
  and~Jianbin~Jiao 
\IEEEcompsocitemizethanks{
\IEEEcompsocthanksitem W. Deng and L. Zheng are with the Research School of Computer Science, Australian National University, CBR, Australia. \protect\\
E-mail: dengwj16@gmail.com, liangzheng06@gmail.com 
\IEEEcompsocthanksitem Y. Yang is with Centre for Artificial Intelligence, University of Technology Sydney, NSW, Australia. \protect\\
E-mail: yi.yang@uts.edu.au
\IEEEcompsocthanksitem Q. Ye, and J. Jiao are with the University of Chinese Academy of Sciences, Beijing, China.\protect\\
E-mail: qxye@ucas.ac.cn, jiaojb@ucas.ac.cn.
\IEEEcompsocthanksitem Corresponding authors: J. Jiao and L. Zheng.}
}

\IEEEtitleabstractindextext{%
\begin{abstract}
\textcolor{black}{Person re-identification (re-ID) models often fail to generalize well to new domains. We propose a ``learning via translation” framework based on the Generative Adversarial Network (GAN). It consists of two components, \ie, 1) translating the labeled source images to style of the target domain, and 2) learning a re-ID model for testing on the target domain using the translated images. Typically, source-target translation suffers from information loss with respect to the discriminative cues that form human identity. To this end, we propose a similarity-preserving generative adversarial network (SPGAN) and its upgraded version, end-to-end SPGAN (eSPGAN).} 
SPGAG improves the first component of the framework. It enforces two heuristic constraints in an unsupervised manner, 1) preserving self-similarity of human identity, and 2) introducing domain dissimilarity, such that the source images preserve the discriminative cues while being transferred to the target style. In comparison, eSPGAN seamlessly integrates the two components of the framework. During its end-to-end training, feature learning guides image translation to preserve the underlying identity information of an image. Meanwhile, image translation improves feature learning by providing identity-preserving training samples of the target domain style. Experiment on two large-scale datasets shows that both SPGAN and eSPGAN obtain state-of-the-art domain adaptation results.

\end{abstract}

}

\maketitle

\IEEEdisplaynontitleabstractindextext
\IEEEpeerreviewmaketitle

\IEEEraisesectionheading{\section{Introduction}\label{sec:introduction}}
\IEEEPARstart{T}{his} article studies the domain adaptation problem in person re-ID under a ``learning via translation" framework. In our setting, the source domain is fully annotated with identity labels, and the target domain does not have any ID labels. In the community, domain adaptation of re-ID is gaining increasing popularity, because of 1) the expensive labeling process and 2) 
when models trained on one dataset are directly used on another, the re-ID accuracy drops dramatically \cite{fan17unsupervised} due to dataset bias \cite{DBLP:conf/cvpr/TorralbaE11}.

A commonly used strategy to above-mentioned problems is unsupervised domain adaptation (UDA). But this line of methods usually assumes that the source and target domains contain the same set of classes. This assumption does not hold in person re-ID because different re-ID datasets usually contain entirely different persons (classes). 
In UDA, a recent trend is image-level domain translation \cite{ hoffman2018cycada, DBLP:journals/corr/BousmalisSDEK16, DBLP:conf/nips/LiuT16}, which motivates us to explore a ``learning via translation'' framework. The framework consists of two components. First, labeled images from the source domain are translated to the target domain, so the translated images and images from the target domain share similar styles, \eg, backgrounds, resolutions, and light conditions. Second, the style-translated images and their associated labels are used for supervised learning in the target domain. In literature, commonly used image-level translation methods include \cite{DiscoGAN, DualGAN, cycle,stargan}. In our work, we adopt CycleGAN in the baseline. 

\begin{figure*}[t]
\setlength{\abovecaptionskip}{0cm}
\setlength{\belowcaptionskip}{0cm}
\begin{center}
\includegraphics[width=0.85\linewidth]{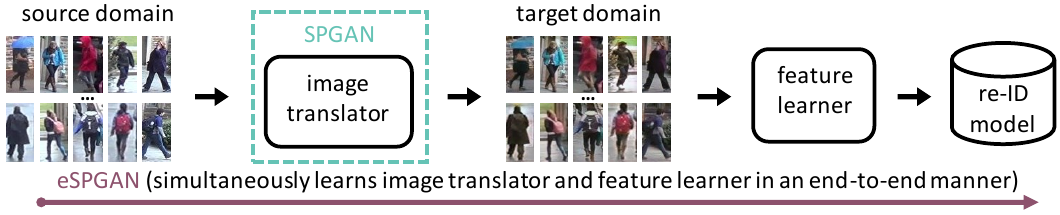}\end{center}
\caption{Pipeline of the ``learning via translation'' framework. First, we translate the labeled images from a source domain to a target domain. Second, we train re-ID models with the translated images using supervised feature learning methods. \textcolor{black}{SPGAN is only used to improve the first component of the framework, while eSPGAN simultaneously learns the two components in an end-to-end-manner.} }
\label{fig:framework}
\end{figure*}
\textcolor{black}{In person re-ID, there is a distinct yet unconsidered requirement for the baseline described above: the visual content associated with the ID label of an image should be preserved during image-image translation. In our scenario, such visual content usually refers to the underlying (latent) ID information for a foreground pedestrian.
To meet this requirement tailored for re-ID, we first propose a heuristic solution, named Similarity Preserving Generative Adversarial Network (SPGAN). Then, we further study the relation between feature learning and image translation, and propose eSPGAN, an upgrade version of SPGAN.}

SPGAN is motivated by two aspects. First, a translated image, despite of its style changes, should contain the same underlying identity with its corresponding source image. Second, in re-ID, the source and target domains contain two entirely different sets of identities. Therefore, a translated image should be different from any image in the target dataset in terms of the underlying ID. SPGAN is composed of an Siamese network (SiaNet) and a CycleGAN. Using a contrastive loss, the SiaNet pulls close a translated image and its counterpart in the source, and push away the translated image and any image in the target. In this manner, the contrastive loss satisfies the specific requirement in re-ID. Note that, the added constraints are unsupervised, \emph{i.e.,} the source labels are not used in source-target image translation.
Through the coordination between CycleGAN and SiaNet, we are able to generate samples which not only possess the style of target domain but also preserve their underlying ID information from the source domain. 

Essentially, SPGAN focuses only on improving the first component of the ``learning via translation" framework, \ie, source-target image translation, which is actually independent of the feature learning component. Thus, the impact of image translation on feature learning and the reverse remains unknown. 
A natural question then arises: can these two components be jointly optimized, so that they could benefit each other? 

\textcolor{black}{In light of this question, we propose eSPGAN by seamlessly integrating the two components into an end-to-end training system. 
In eSPGAN, there exists a mutually beneficial interactive loop between image translation and re-ID feature learning. Thus, the translated images are better suited for the re-ID task, leading to higher re-ID accuracy.}
More specifically, feature learning guides image translation to preserve the identity of images during translation; in return,  image translation delivers the knowledge of how a person looks like on the target domain to feature learning. During training, we alternately optimize the two components, so that knowledge and constraint of both components are gradually transferred to each other. 

\textcolor{black}{This paper extends our previous conference paper \cite{image-image18} in several aspects. Primarily, we integrate the two components of ``learning via translation” framework into an end-to-end system, yielding eSPGAN. In eSPGAN, we discover the mutually benefit between image translation and re-ID feature learning. In addition, insightful analyses of the visual changes conducted by the image translation are provided. Also, the difference from other similarity-preserving generation methods is discussed. Finally, we present significant extensions in the experiment to validate the effectiveness of our methods: 1) we report higher results of baseline methods and SPGAN with our latest implementations; 2) we extensively investigate eSPGAN. }

\textcolor{black}{Overall, the contributions of this study are mainly in the following four aspects:
\begin{itemize}
\item To address the domain adaptation in person re-ID, we present a ``learning via translation” framework. We further introduce SPGAN, a heuristic method, to preserve the underlying ID information during source-target image translation. SPGAN better qualifies the translated images and produces competitive domain adaptation accuracy.
\item We report the mutual benefit between generative image translation and discriminative feature learning. Inspired by this, we propose ePSGAN, an upgraded version of SPGAN, by simultaneously optimizing image translation and feature learning for the domain adaptative person re-ID. In eSPGAN, there exists a beneficial interactive loop between image translation and re-ID feature learning. Thus, the translated images are better suited for re-ID feature learning, leading to higher re-ID accuracy.
\item We provided insightful analyses of the ``style” change introduced by image translation. We find that ``style” change involves various factors, such as illumination and color composition. This helps us take a closer look at the ``style” transfer and gives a better understanding of the dataset bias.
\item As a minor contribution, we propose a local max pooling (LMP) scheme as a post-processing step. LMP is tailored for the domain adaptation scenario, and consistently improves over SPGAN and eSPGAN.
\end{itemize}}



The reminder of this paper is organized as follows. Related work is presented in Section \ref{sec:Related Work}. Section \ref{sec:Proposed Method} describes SPGAN and eSPGAN. In Section \ref{sec:experiments} , the experimental results are presented and analyzed. Section \ref{sec:Conclusion and future work} concludes the paper.

\section{Related Work}\label{sec:Related Work}
\textbf{Image-image translation.} 
Image-image translation aims at learning a mapping function between two domains. 
As a representative image-image translation method, the ``pixel2pixel" framework uses input-output pairs for learning a mapping from input to output images.
In practice, the paired training data is often difficult to acquire and hence the unpaired image-image translation is often more applicable.
 To tackle the unpaired setting, a cycle consistency loss is introduced by DiscoGAN \cite{DiscoGAN}, DualGAN \cite{DualGAN}, and CycleGAN \cite{cycle}. Benaim \etal \cite{DBLP:journals/corr/BenaimW17} propose an unsupervised distance loss for one side domain mapping. 
Liu \etal \cite{DBLP:journals/corr/LiuBK17}  propose a general framework by making a shared latent space assumption that the corresponding images in two domains are mapped to the same latent code. 
Recently, some methods \cite{stargan, munit} have been proposed to learn the relations among multiple domains.
In this work, while we aim to find mapping functions between the source domain and target domain, our primary focus is similarity-preserving mapping.

Neural style transfer \cite{perceptual, DBLP:conf/eccv/LiW16, DBLP:conf/icml/UlyanovLVL16, DBLP:journals/corr/ChenS16f, DBLP:journals/corr/LiFYWL017, DBLP:journals/corr/HuangB17, DBLP:conf/ijcai/LiWLH17} is another strategy of image-image translation, which aims at rendering the content of an image in the style of another image.
Gatys \etal \cite{GatysEB16} employ an optimization process to match feature statistics in layers of a convolutional network. The optimization is replaced by a feed-forward neural network in \cite{perceptual, DBLP:conf/icml/UlyanovLVL16,DBLP:conf/eccv/LiW16}. Huang \etal \cite{DBLP:journals/corr/HuangB17} propose a AdaIN layer for arbitrary style transfer. Unlike the neural style transfer, our work focuses on learning the mapping function between two domains, rather than two images. 

\textbf{Unsupervised domain adaptation.}
Our work is related to unsupervised domain adaptation (UDA). Within this community, a portion of methods aim to learn a mapping between source and target distributions \cite{DBLP:conf/eccv/SaenkoKFD10, DBLP:conf/cvpr/GongSSG12,DBLP:conf/iccv/FernandoHST13, DBLP:conf/aaai/SunFS16}. 
As a representative UDA method, Correlation Alignment (CORAL) \cite{DBLP:conf/aaai/SunFS16}  matches the mean and covariance of two distributions.

Other methods seek to find a domain-invariant feature space \cite{DBLP:journals/corr/abs-1709-10190, DBLP:conf/cvpr/LongD0SGY13, DBLP:conf/icml/GaninL15, DBLP:conf/icml/LongC0J15, DBLP:journals/corr/TzengHZSD14, DBLP:journals/jmlr/GaninUAGLLML16, DBLP:journals/corr/AjakanGLLM14}. Long \etal \cite{DBLP:conf/icml/LongC0J15} use the Maximum Mean Discrepancy (MMD) \cite{MMD} for this purpose. Ganin \etal \cite{DBLP:journals/jmlr/GaninUAGLLML16} and Ajakan \etal \cite{DBLP:journals/corr/AjakanGLLM14} introduce a domain confusion loss to learn domain-invariant features. 
In addition, several approaches estimate the labels of unlabeled samples \cite{ChenWB11, RohrbachES13, SenerSSS16,zhang2018collaborative}.  The estimated labels are then used to learn the optimal classifier. Zhang \etal \cite{zhang2018collaborative} propose a progressive method to select a set of pseudo-labeled target samples. Sener \etal \cite{SenerSSS16} use the K-nearest neighbors to predict the labels of target samples.

Recent methods \cite{ hoffman2018cycada, DBLP:journals/corr/BousmalisSDEK16, DBLP:conf/nips/LiuT16} use an adversarial approach to learn a transformation in the pixel space from one domain to another.  The CYCADA \cite{ hoffman2018cycada} maps samples across domains at both pixel level and feature level.
We note that most of the UDA methods assume that class labels are the same across domains. However, the setting in this paper is different, because different re-ID datasets contain entirely different person identities (classes). Therefore, the approaches mentioned above cannot be utilized directly for domain adaptation in person re-ID.

\textbf{Unsupervised person re-ID.}
Unsupervised person re-ID approaches leverage hand-craft features \cite{DBLP:journals/ivc/MaSJ14,DBLP:conf/eccv/GrayT08,DBLP:conf/cvpr/FarenzenaBPMC10,DBLP:conf/cvpr/MatsukawaOSS16,DBLP:conf/cvpr/LiaoHZL15,DBLP:conf/iccv/ZhengSTWWT15} or learning based features  \cite{DBLP:conf/cvpr/ZhaoOW13,DBLP:conf/bmvc/WangGX14} as representation. Hand-craft features can be directly employed in the unsupervised setting, but they do not fully exploit data distribution and fail to perform well on large-scale datasets. Some methods are based on saliency statistics \cite{DBLP:conf/cvpr/ZhaoOW13,DBLP:conf/bmvc/WangGX14}. Yu \etal \cite{CAMEL} use K-means clustering to learn an unsupervised asymmetric metric. 
Peng \etal \cite{DBLP:conf/cvpr/PengXWPGHT16} propose an asymmetric multi-task dictionary learning for cross-data transfer. Wang \etal \cite{wang2018} utilize additional attribute annotations to learn a feature representation space for the unlabeled target dataset. 

Several works focus on label estimation of unlabeled target dataset \cite{DBLP:journals/corr/abs-1709-09297, fan17unsupervised, liu2017stepwise, wu2018exploit}. Fan \etal \cite{fan17unsupervised} propose a progressive method based on the iterations between K-means clustering and IDE \cite{zheng2016mars} fine-tuning. Ye \etal \cite{DBLP:journals/corr/abs-1709-09297} use graph matching for cross-camera label estimation. Liu \emph{et al.} \cite{liu2017stepwise} employ a reciprocal search process to refine the estimated labels. Wu \etal \cite{wu2018exploit} propose a dynamic sampling strategy for one-shot video-based re-ID. 
Our work seeks to learn re-ID models that can be utilized directly on the target domain and can potentially cooperate with label estimation methods in the model initialization.

Recently, some Generative Adversarial Network (GAN) based methods are applied to explore domain adaptive re-ID models. 
The most recent HHL approach \cite{zhong2018generalizing} enforces cameras invariance and domain connectedness simultaneously for learning more generalizable embeddings on the target domain.
PTGAN \cite{ptgan}, a concurrent work, adopts CycleGAN \cite{cycle} to generate training samples on the target domain. The common characteristic of PTGAN and our SPGAN lies in that they both consider the similarity between the generated and original image. The key difference is that PTGAN requires the foreground mask using an extra segmentation step, while SPGAN leverages two unsupervised heuristic constraints to preserve the identity of translated images.

\section{Proposed Method}\label{sec:Proposed Method}
For unsupervised domain adaptation in person re-ID, we are provided with  an annotated dataset $\mathcal{S}=\{(x_{i}^{s}, y_{i}^{s})\}_{i=1}^{n_{s}}$ of $n_{s}$ labeled images associated with $|\mathcal{C}_{s}|$ identities from the source domain and an unlabeled dataset $\mathcal{T}=\{x_{i}^{t}\}_{i=1}^{n_{t}}$ of $n_{t}$ unlabeled images associated with $|\mathcal{C}_{t}|$ identities from the target domain. Note that the label space of the source domain $\mathcal{C}_{s}$ is totally different from that in the target domain $\mathcal{C}_{t}$, \ie,  $\mathcal{C}_{s} \cap \mathcal{C}_{t} = \emptyset$. The goal of this paper is to use both the labeled source images and the unlabeled target images to train a re-ID model that generalizes well on the target domain.
Briefly, in Section \ref{framework}, we introduce the ``learning via translation"  framework. In Section \ref{SPGAN}, we revisit SPGAN. In Section \ref{eSPGAN}, we extend the SPGAN to eSPGAN to comprehensively study the relation between source-target image translation and feature learning.

\subsection{Learning via Translation} \label{framework}
The ``learning via translation'' framework shown in Fig. \ref{fig:framework}. This framework consists of two components, \emph{i.e.,} source-target image translation for training data creation, and supervised feature learning for person re-ID. 

\begin{itemize}
\setlength{\itemsep}{-0pt}
\item \textbf{Source-target image translation.} Using a generative function $G(\cdot)$ that translates the annotated dataset $\mathcal{S}$ from the source domain to target domain in an unsupervised manner, we ``create'' a labeled training dataset $G(\mathcal{S})$ on the target domain.

\item \textbf{Feature learning.} With the translated dataset $G(\mathcal{S})$ that contains labels, supervised feature learning methods can be applied to train re-ID models. 
\end{itemize}

\textbf{In the baseline}, source-target image translation is achieved by CycleGAN. For feature learning, we adopt several existing methods, such as identity discriminative embedding (IDE+) \cite{zheng2016mars} and part-based convolutional baseline (PCB) \cite{PCB}.

As analyzed in Section \ref{sec:introduction}, we aim to preserve the ID-related cues for each translated image. We emphasize that the ID information should not be the background or image style, but should be underlying and latent. 
To this end, \textbf{SPGAN} focuses on improving the image translation component of Fig. \ref{fig:framework}, so as to improve the re-ID accuracy. \textbf{eSPGAN} integrates image translation and feature learning into an end-to-end training system, and yields higher re-ID accuracy.
\begin{figure}[t]
\setlength{\abovecaptionskip}{0cm}
\setlength{\belowcaptionskip}{0cm}
\begin{center}
\includegraphics[width=1 \linewidth]{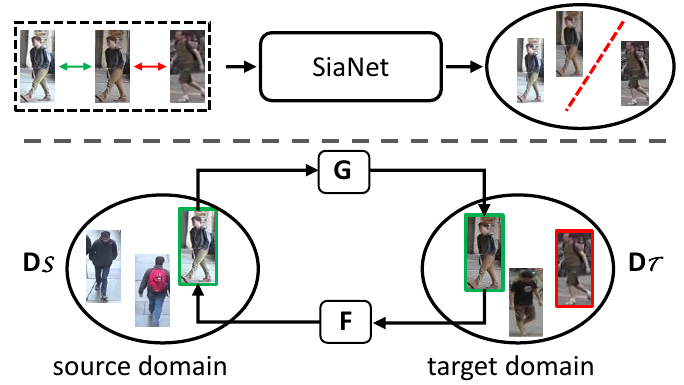}
\end{center}
\caption{SPGAN consists of two components: SiaNet (top) and CycleGAN (bottom). CycleGAN learns mapping functions $G$ and $F$ between the two domains, and SiaNet constraints the learning of mapping functions using two heuristic similarity-preserving losses.}
\label{fig:SPGAN}
\end{figure}

\subsection{SPGAN} \label{SPGAN}
SPGAN mainly consists of SiaNet and CycleGAN, as shown in Fig. \ref{fig:SPGAN}. During the training, CyleGAN aims to learn mapping functions between source and target domains, and SiaNet learns a latent space that constrains the learning of mapping functions. 

\textcolor{black}{\textbf{CycleGAN} introduces two generator-discriminator pairs, $\{G, D_{\mathcal{T}}\}$ and $\{F, D_{\mathcal{S}}\}$, which map a sample from source (target) domain to target (source) domain and produce a sample that is indistinguishable from those in the target (source) domain, respectively.
The overall objective of CycleGAN can be written as,
\begin{small}
\begin{equation}
\begin{split}
\mathcal{L}_{cyc}(G, F, D_{\mathcal{T} }, D_{\mathcal{S}})=& \mathcal{L}_{adv}(G, D_{\mathcal{T}}) + \mathcal{L}_{adv}(F, D_{\mathcal{S}}) \\
&+ \alpha \mathcal{L}_{rec}(G, F) ,
\end{split}
\label{cycleloss}
\end{equation}
\end{small}
where ${L}_{adv}$ denotes the standard adversarial loss \cite{GAN}, ${L}_{rec}$ represents the cycle consistency loss \cite{cycle}, and
$\alpha$ controls the relative importance of the cycle-consistent loss. We would like to refer the readers to the CycleGAN \cite{cycle} for more details about these loss functions.}

\textcolor{black}{In the experiment, we observe in Fig. \ref{samples} (b) that the model may change the color composition of the input image. This is undesirable for re-ID feature learning. Thus, we introduce the inside-domain identity constraint \cite{DBLP:journals/corr/TaigmanPW16} as an auxiliary for image translation. Inside-domain identity constraint is introduced to regularize the generator to be an identity matrix on samples from the expected domain, written as:
\begin{equation}
\begin{split}
\mathcal{L}_{ide}(G, F) = &\mathbb{E}_{x^{s} \sim p_{\text{data}(\mathcal{S})}}{\Vert F(x^{s}) - x^{s}\Vert}_{1}\\
&+ \mathbb{E}_{x^{t} \sim p_{\text{data}(\mathcal{T})}}{\Vert G(x^{t}) - x^{t}\Vert}_{1},
\end{split}
\end{equation}
\label{eq:Identity} 
where $ p_{\text{data}(\mathcal{S})}$ and $ p_{\text{data}(\mathcal{T})}$ denote the sample distributions in the source and target domain, respectively.}

\textbf{Similarity preserving loss function.}
We utilize the contrastive loss \cite{DBLP:conf/cvpr/HadsellCL06} to train the SiaNet $M$,
\begin{equation}
\begin{split}
\mathcal{L}_{con}(i, x_{1}, x_{2}) = &(1-i)\{\max(0, m - d)\}^2 + id^2,
\end{split}
\label{eq:Contrastive}
\end{equation}
where $x_{1}$ and $x_{2}$ form a pair of input vectors, $d$ denotes the Euclidean distance between the normalized embeddings of the two input vectors, and $i$ represents the binary label of the pair. $i= 1$ if $x_{1}$ and $x_{2}$ are a positive pair; $i= 0$ if $x_{1}$ and $x_{2}$ are a negative pair. $m \in [0, 2]$ is the margin that defines the separability of the negative pair in the embedding space.
When $m=0$, the loss of the negative training pair is not backpropagated in the system. When $m > 0$, both positive and negative sample pairs are considered. A larger $m$ means the loss of negative training samples has a higher impact in back-propagation. 

\begin{figure}[t]
\setlength{\abovecaptionskip}{-0.1cm} 
\setlength{\belowcaptionskip}{-0.2cm}
\begin{center}
\includegraphics[width=0.8 \linewidth]{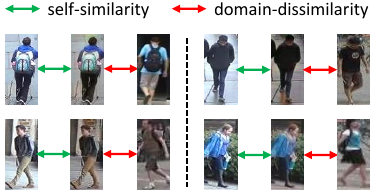}
\end{center}
\caption{Illustration of self-similarity and domain-dissimilarity. In each triplet, left: a source-domain image, middle: a source-target translated version of the source image, right: an arbitrary target-domain image. We require that 1) a source image and its translated image should contain the same ID, \ie, self-similarity, and 2) the translated image should be of a different ID with any target image, \ie, domain dissimilarity. Note: the source and target domains contain entirely different IDs. Best viewed in color. }
\label{fig:carton}
\end{figure}

\emph{Training data construction.} In Eq. \ref{eq:Contrastive}, the contrastive loss uses binary labels of input image pairs. In this article, we design these image pairs to reflect the proposed ``self-similarity'' and ``domain-dissimilarity'' principles. Note that, \emph{training pairs are constructed in an unsupervised manner}, so that we use the contrastive loss without additional annotations.

\begin{itemize}
\item \emph{self similarity.} Suppose two samples denoted as $x^{s}$ and $x^{t}$ come from the source domain and target domain, respectively. Given $G$ and $F$, we define two positive pairs: 1) $x^{s}$ and $G(x^{s})$, 2) $x^{t}$ and $F(x^{t})$. In either image pair, the two images contain the same person; the only difference is that they have different styles. In the learning procedure, we encourage SiaNet M to pull these two images close. 
\item \emph{domain dissimilarity.} For generators $G$ and $F$, we also define two types of  negative training pairs: 1) $G(x^{s})$ and $x^{t}$, 2) $F(x^{t})$ and $x^{s}$. This design of negative training pairs is based on the prior knowledge that datasets in different re-ID domains have entirely different sets of IDs. Thus, a translated image should be of a different ID from any target image. In this manner, the network M pushes two dissimilar images away. Training pairs are shown in Fig. \ref{fig:carton}. Some positive pairs are also shown in (a) and (d) of each column in Fig. \ref{samples}.
\end{itemize}



\textbf{Overall objective of SPGAN.} The overall objective function of SPGAN can be written as,
\begin{equation}
\begin{split}
\mathcal{L}_{sp}(G, F, D_{\mathcal{T} }, D_{\mathcal{S}}, M)= & \mathcal{L}_{cyc}(G, F, D_{\mathcal{T} }, D_{\mathcal{S}})\\
&+\beta \mathcal{L}_{ide}(G, F) \\
&+\gamma \mathcal{L}_{con}(G, F, M),
\end{split}
\label{spgan_Objective}
\end{equation}
where the first two losses belong to the CycleGAN formulation \cite{cycle}, the parameters $\beta$ and $\gamma$ control the relative importance of the identity loss of CycleGAN and the proposed contrastive constraint. In other words, the contrastive loss induced by SiaNet imposes a new constraint on the GAN system. The optimization process of SPGAN is,
\begin{equation}
 G^{*}, F^{*}, M^{*}=\mathop{\arg}\mathop{\min}_{G, F, M} \mathop{\max}_{D_{\mathcal{T} }, D_{\mathcal{S}}}\mathcal{L}_{sp}(G, F, D_{\mathcal{T} }, D_{\mathcal{S}}, M).
\label{SPGAN Full Objective}
\end{equation}

\textcolor{black}{\emph{Training procedure of SPGAN.} In practice, we replace the standard adversarial loss in Eq. \ref{cycleloss} by the least-squares loss \cite{cycle,Mao_2017_ICCV} to make training more stable.
Specifically, for the adversarial loss
$\mathcal{L}_{adv}(G, D_{\mathcal{T}})$ in Eq. \ref{cycleloss}, we train the $G$ to minimize $\mathbb{E}_{x^{s} \sim p_{\text{data}(\mathcal{S})}}[D_{\mathcal{T}}(G(x^{s})-1)^{2}]$ and train the $D_{\mathcal{T}}$ to minimize 
$\mathbb{E}_{x^{s} \sim p_{\text{data}(\mathcal{S})}}[D_{\mathcal{T}}(G(x^{s}))^{2}]
+ \mathbb{E}_{x^{t} \sim p_{\text{data}(\mathcal{T})}}[D_{\mathcal{T}}(G(x^{t})-1)^{2}]$.}

There are three parts in SPGAN, generators, discriminators, and SiaNet. They are optimized alternately during training. When the parameters of any one part are updated, the parameters of the remaining two parts are fixed. We train SPGAN until  convergence or reaching maximum iterations.

\subsection{Feature Learning}\label{sec:feature_learning}
Feature learning is the second component of the ``learning via translation'' framework. Once we have the style-transferred dataset $G(\mathcal{S})$ composed of translated images and their associated labels, the feature learning step is the same as supervised methods. We adopt the baseline ID-discriminative Embedding (IDE+) following the practice in \cite{zheng2016mars, zhong2018camera}.  Given an annotated dataset $G(\mathcal{S})$, IDE+ aims to learn a model $C$ by $| \mathcal{C}_{s}|$-way classification with a cross-entropy loss. This corresponds to, 
 \begin{equation}
\begin{split}
\mathcal{L}_{c}(C) = -\mathbb{E}(G(x^{s}), y^{s})  \sum_{k=1}^{|\mathcal{C}_{s}|} \mathbbm{1}_{[k = y^{s}]}  \log \left( \sigma(C^{(k)}(G(x^{s}))) \right),
\end{split}
\label{loss:cross}
 \end{equation}
where $\sigma$ denotes the softmax activation function.

\subsection{End-to-end SPGAN (eSPGAN)} \label{eSPGAN}
SPGAN only focuses on preserving the image identity information during image translation. It is independent of the subsequent feature learning component of ``learning via translation'' framework. We believe the two components of the framework could benefit each other if jointly trained: 1) feature learning could guide image translation to generate identity-preserving images without heuristic constraints; 2) a stronger image translator will generate more beneficial samples for feature learning, leading to more robust person descriptors  for the target domain. 
To this end, this article further studies the inherent relation between these two components. Specifically, we propose eSPGAN by merging the two components into an end-to-end training system. 

\subsubsection {Objective}\label{espgan}
eSPGAN is a unified system. It translates images to the target domain and learns re-ID features simultaneously. 
Following the idea of learning via translation, eSPGAN consists of two models: an image translator and a feature learner (Fig. \ref{fig:framework}). The image translator translates source images to the style of the target domain, and the feature learner learns discriminative embeddings that can be used on the target domain. Note that feature learner is differentiable with respect to the elements in the translated image $G(x^{s})$. Thus, the whole system can be trained end-to-end.

\textbf{Overall objective of eSPGAN.}
On the top of CycleGAN, we adopt the feature learner as the supervisor of the image translation. We alternately optimize the feature learner and the image translator, 1) when training the feature learner, we keep the image translator fixed, and learn a model $C$ by $| \mathcal{C}_{s}|$-way classification; 2) when training the image translator, we keep the feature learner fixed, and use its re-ID accuracy as the guidance. The feature learner will propagate a supervision signal (Eq. \ref{loss:cross}) to update the image translator, so that the translated images could be classified correctly by the former. Namely, the visual content associated with the identity information of an image is preserved. 
The overall objective function of eSPGAN can be written as,
\begin{equation}
\begin{split}
\mathcal{L}_{esp}(G, F, D_{\mathcal{T} }, D_{\mathcal{S}}, C)= & \mathcal{L}_{cyc}(G, F, D_{\mathcal{T} }, D_{\mathcal{S}})\\
&+\beta \mathcal{L}_{ide}(G, F) \\
&+\lambda \mathcal{L}_{c}(G, C),
\end{split}
\label{eSPGAN_Objective}
\end{equation}
where the first two losses belong to the CycleGAN formulation \cite{cycle}. The parameter $\lambda$ controls the relative importance of the feature learner constraint $\mathcal{L}_{c}(C)$. The optimization process of eSPGAN is,
\begin{equation}
 G^{*}, F^{*}, C^{*}=\mathop{\arg}\mathop{\min}_{G, F, C} \mathop{\max}_{D_{\mathcal{T} }, D_{\mathcal{S}}}\mathcal{L}_{esp}(G, F, D_{\mathcal{T} }, D_{\mathcal{S}}, C).
\label{SPGAN Full Objective}
\end{equation}

\begin{figure}[t]
\setlength{\abovecaptionskip}{0cm} 
\setlength{\belowcaptionskip}{0cm}
\begin{center}
\includegraphics[width=0.85\linewidth]{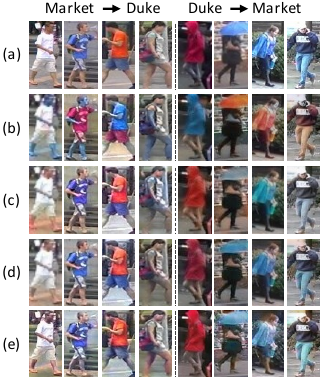}
\end{center}
\caption{Visual examples of image-image translation. The left four columns map Market images to the Duke style, and the right four columns map Duke images to the Market style. From top to bottom: (a) original image, (b) output of CycleGAN, (c) output of CycleGAN + $L_{ide}$, (d) output of SPGAN, and (e) output of eSPGAN, respectively. 
\textcolor{black}{We observe some visual changes after image translation, such as resolution, illumination, color, and background.
Moreover, SPGAN and eSPGAN have the characteristic of preserving underlying semantics of input images. 
Thus, their translated images will share some visual similarities with the original images. 
Best viewed in color.}}
\label{samples}
\end{figure}

\textbf{Training procedure of eSPGAN.} 
\textcolor{black}{Similar to SPGAN, we also use the least-squares loss \cite{cycle,Mao_2017_ICCV} for training the generator-discriminator pairs. 
The proposed eSPGAN adopts an alternate optimization procedure. 
There are three parts in eSPGAN: generators, discriminators and feature learner (IDE+). While updating the parameters of a single part, the parameters of the other two parts are fixed. We train eSPGAN until convergence or the maximum iterations is reached.} 

\subsubsection{Discussions on eSPGAN} \label{know}
\textcolor{black}{In this section, we present a comprehensive discussion on  eSPGAN.  
 We first introduce the knowledge transfer mechanism of eSPGAN. Then, we analyze the crucial factor that determines whether the end-to-end system can work effectively. Moreover, the difference from other similarity-preserving methods is provided. Finally, we fully compare ePSGAN and SPGAN.}

\begin{figure}[t]
\setlength{\abovecaptionskip}{0cm} 
\setlength{\belowcaptionskip}{0cm}
\begin{center}
\includegraphics[width=1\linewidth]{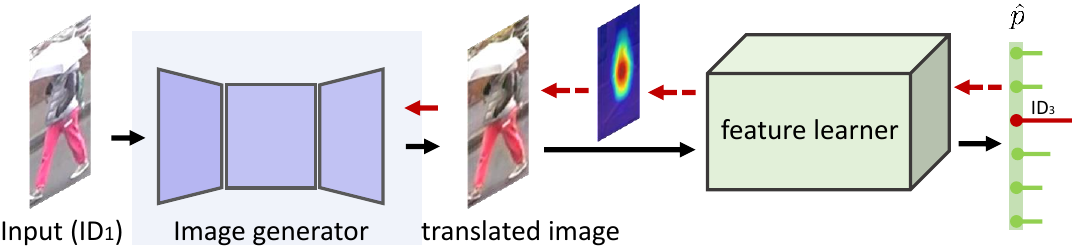}
\end{center}
\caption{Illustration of transferring knowledge of the person identity from the feature learner to the image translator. In this example, the identity of the original image is $\text{ID}_1$, but its corresponding translated image is miss-classified to $\text{ID}_3$ by the feature learner. Namely, the identity similarity between the original image and its translated image is not preserved. To solve this problem, feature learner directly backpropagates the gradients to the input pixels of the translated image, and further updates the image translator (the red arrow). Thus, the image translator is guided to preserve person identity during translation. Note that the feature learner is fixed when we train the image translator.}
\label{fig:KT1}
\end{figure}

\textbf{1. Bidirectional knowledge transfer.} The optimizing procedure of eSPGAN can be regarded as transferring knowledge between the two components. The knowledge transfer is bidirectional: the feature learner tells the image translator \emph{how to preserve the identity of an image}; the image translator provides \emph{what a person from source domain looks like in target domain} for the feature learner.

\emph{(a) Feature learner guides image translator.} Feature learner has the ability to distinguish between different identities, so it serves as a guide for image translator. During training, the translated image passes through the feature learner with fixed parameters and computes the classification loss, corresponding to Eq. \ref{loss:cross}. The feature learner then backpropagates the gradients to the input pixels of the translated image, and further updates the image translator. Thus, the image translator is guided to translate images that benefit the classification of the feature learner. As we can interpret, the translated image preserves its visual content associated with its identity. 

In an example shown in Fig. \ref{fig:KT1}, the translated image is misclassified by the feature learner because its identity is somehow lost during translation. In this case, the feature learner backpropagates a supervision signal to guide the training of the image translator, so that the translated image can be correctly classified.  Namely, the visual content associated with the identity of an image is preserved after image translation.

\begin{figure}[t]
\setlength{\abovecaptionskip}{-0.1cm} 
\setlength{\belowcaptionskip}{0cm}
\begin{center}
\includegraphics[width=0.8 \linewidth]{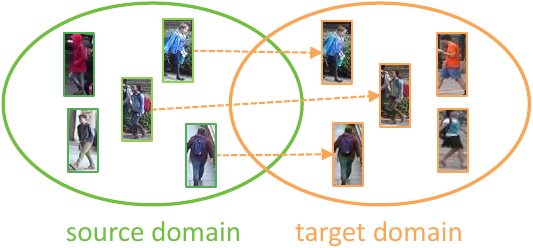}
\end{center}
\caption{Illustration of transferring knowledge of the target domain from image translator to feature learner. Images with green boxes and orange boxes are on the source domain and target domain, respectively. Image translator delivers the knowledge of how a person looks like on the target domain to feature learner. Thus, the feature learner learns a domain invariant feature space by using translated images and source images for training.}
\label{fig:KT2}
\end{figure}

 \emph{(b) Image translator strengthens feature learner.}
As shown in Fig. \ref{fig:KT2}, image translator translates images from the source domain to the target domain, \ie, image translator creates a training dataset with labels in the target domain for feature learner. Based on this, the feature learner can learn discriminative person embeddings for the target domain.
 

\textbf{2. Maintaining the discriminative ability of feature learner.} To provide beneficial knowledge for image translator, feature learner has to maintain its discrimination ability. Several techniques and issues are described below.

\emph{i) Pre-training the feature learner on the source dataset.} 
\textcolor{black}{The feature learner adopts ResNet-50 \cite{DBLP:conf/cvpr/HeZRS16} pre-trained on ImageNet \cite{imagenet} as the backbone. We first fine-tune the feature learner on the labeled source dataset $\mathcal{S}$, such that it has discriminative ability at the beginning of eSPGAN training}. 

\emph{ii) Real data regularization.}
There exist some poorly translated images, especially at the early epochs of eSPGAN training. By ``poorly translated image'', we mean two types of images. First, the image translator fails to generate high-quality images from the source to the target domain. Second, the identity of a translated image is largely lost. These poorly translated images are likely to be misclassified by feature learner and produce relatively large losses. This might cause the instability problem that affects the learning of eSPGAN.

\textcolor{black}{To this end, 
We also use the source images when training eSPGAN. In practice, besides the batch of translated source images, we sample another batch of unaltered source images for training the feature learner.}
This practice guarantees that the feature learner will not be led to divergence by the poorly translated images.
Moreover, using both the source and the translated images allows feature learner to learn domain invariant person embeddings. Namely, the learned feature is effective for both the source and target domains.

In late training epochs, the image translator has the ability to improve the poorly translated images based on the gradient computed by feature learner. Thus, the translated images usually have high quality and largely preserve person identities. At this stage, their effectiveness for learning desirable feature at the target domain is understandable.

\textbf{3. Different from other similarity-preserving methods.} 
There are some existing methods that also focus on the similarity-preserving property of generated images \cite{hoffman2018cycada, liu2018pose, chen2018zero, ShiriYPHK19}. For example, CYCADA \cite{ hoffman2018cycada} and Pose-transfer \cite{liu2018pose} both propose to utilize a model that is pre-trained on real images to preserve the semantics of generated images. SP-AEN \cite{chen2018zero} uses pre-trained AlexNet \cite{alexnet2012} to preserve perceptual information of an generated image. SRN \cite{ShiriYPHK19} also adopts pre-trained FaceNet model to maintain the identity of the recovered face image.
These existing methods all keep the pre-trained model fixed, \ie, the parameters of the pre-trained model are not updated during training. Under this case, these methods can be viewed as the content loss \cite{perceptual} in the style transfer.

Departing from these methods, \textbf{we actually find that pre-trained feature learner should be updated during training.} 
We speculate the reason is that pre-trained feature learner only contains the semantic knowledge about the source domain, and it is not effective in classifying target-style images. As a consequence, a translated image might still be mis-classified by the pre-trained model even if it has successfully preserved its identity during translation.

\textcolor{black}{In the person re-ID community, there are two end-to-end methods \cite{wang2018cascaded, jiao2018deep} for the low-resolution re-ID task. Our work is inherently different from both works in several essential aspects. 
First, in CSR-GAN \cite{wang2018cascaded}, the loss of re-ID does not influence the super-resolution network. In comparison, in eSPGAN the re-ID model and the image translator are well-aligned and have the impact on each other. 
Second, super-solution network in SING \cite{jiao2018deep} and image translator have the substantial difference in loss function and network structure. For example, SING learns to do super resolution with ground truths in the form of low-resolution and high-resolution pairs. However, the image translator does not have paired data, and it learns to map domains from data distributions in an unsupervised manner. Thus, eSPGAN and SING \cite{jiao2018deep} have significant different working mechanisms, making them completely different end-to-end systems.}


%

\textcolor{black}{\textbf{4. eSPGAN vs. SPGAN.} 
To understand the differences between SPGAN and eSPGAN, we thoroughly compare them in three aspects:}

\textcolor{black}{i) Network architecture. SPGAN only focuses on learning an image translator, and the re-ID feature is separately learned. In comparison, eSPGAN consists of both the image translator and the re-ID feature learner and learns them in an end-to-end manner. Thus, SPGAN is not end-to-end trainable, but eSPGAN is.}

\textcolor{black}{ii) Working mechanism. Both aiming at similarity-preserving image translation, SPGAN enforces this property by two unsupervised heuristic constraints, while eSPGAN does so by optimally facilitating the re-ID model learning. eSPGAN seamlessly integrates image translation and re-ID model learning, which allows us to gradually leverage the knowledge of the two components to learn discriminative embeddings for the target domain. In the experiment, we also applied the two unsupervised heuristic constraints to eSPGAN, but this does not bring any improvement.}

\textcolor{black}{iii) Training procedure. The alternative training procedures of eSPGAN and SPGAN appear similar from a high-level perspective. However, they use significantly different loss functions and architectures, and as such their training logistics are significantly different.}

\subsection{Local Max Pooling} 
After describing SPGAN (Section \ref{SPGAN}) and eSPGAN (Section \ref{eSPGAN}), this article also introduces a useful technique for person re-ID under the domain adaptation setting, named local max pooling (LMP). LMP is not used in training; it works on a well-trained re-ID model, and is used for feature extraction of the query and gallery images.
This method can reduce the impact of noisy signals incurred by fake translated images. 

\begin{figure}[t]
\setlength{\abovecaptionskip}{-0.1cm} 
\setlength{\belowcaptionskip}{-0.2cm}
\begin{center}
\includegraphics[width=0.97 \linewidth]{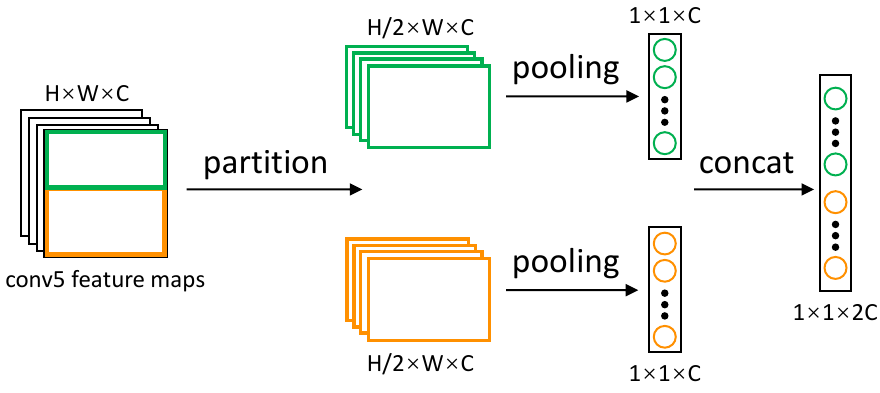}
\end{center}
\caption{Illustration of LMP. We partition the feature map into $P$ ($P=2$ in this example) parts horizontally. We conduct global max/ average pooling on each part and concatenate the resulting feature vectors as the final representation.}
\label{fig:LMP}
\end{figure}

Specifically, in the original ResNet-50, global average pooling (GAP) is conducted on the last Convolution layer (Conv5). In the LMP (Fig. \ref{fig:LMP}), 
we first partition the Conv5 feature maps to $P$ horizontal parts, and then conduct global max pooling (GMP) or global average pooling (GAP) on each part. Finally, we concatenate the output of GMP or GAP of each horizontal part as the final feature representation. This procedure is non-parametric, and can be directly used in the testing phase. In the experiment, we will compare local max pooling and local average pooling, and demonstrate the superiority of the former. Moreover, we will show that LMP is useful under the domain adaptation setting and does not yield improvement under the normal setting where training and testing are conducted on the same domain.

\section{ Experimental evaluation} \label{sec:experiments}
\subsection{Datasets}
We evaluate the proposed methods on two large-scale datasets, \ie, Market-1501 \cite{DBLP:conf/iccv/ZhengSTWWT15} and DukeMTMC-reID \cite{zheng2017unlabeled}, 
and investigate the components of our methods in details. 

\textbf{DukeMTMC-reID} consists of 34,183 bounding boxes of 1,404 identities. There are 16,522  images from 702 identities for training, 2,228 query images from another 702 identities and 17,661 gallery images for testing.  Each identity is captured by at most 8 cameras. DukeMTMC-reID is denoted as Duke for short.

\begin{figure}[t]
\setlength{\abovecaptionskip}{0cm} 
\setlength{\belowcaptionskip}{0cm}
\begin{center}
\includegraphics[width=1\linewidth]{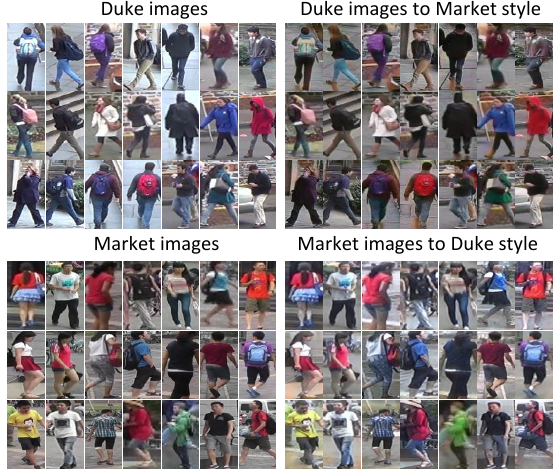}
\end{center}
\caption{Sample images of (upper left:) DukeMTMC-reID dataset, (lower left:) Market-1501 dataset, (upper right:) Duke images which are translated to Market style, and (lower right:) Market images translated to Duke style. \textcolor{black}{We use SPGAN for the source-target image translation. We observe that image resolution, illumination, color, and background are changed.} Best viewed in color.}
\label{fig:sample_images}
\end{figure}

\textbf{Market-1501} contains 12,936 training images and 19,732 gallery images (with 2,793 distractors) detected by DPM\cite{felzenszwalb2008discriminatively}.  It is split into 751 identities for training and 750 identities for testing. There are 3,368 hand-drawn bounding boxes from the 750 identities used as queries. Each identity is captured by at most 6 cameras. We also denote Market-1501 as Market for short.
Sample images of two datasets are shown in Fig. \ref{fig:sample_images}.

\textbf{Evaluation protocol.} We adopt rank-1 accuracy for re-ID evaluation, which counts the percentage of queries that successfully retrieve a true match at rank 1. 
Besides, since multiple true positives should be returned for each query bounding box, we adopt the mean average precision (mAP) for re-ID evaluation. For Market and Duke, we use the evaluation packages provided by the \cite{DBLP:conf/iccv/ZhengSTWWT15} and \cite{zheng2017unlabeled}. 
If not specified, the re-ID results in this paper are reported under the single-query setting.

\subsection{Implementation Details} \label{implementation detail}
\textbf{Feature learning method.} To learn the re-ID model, we adopt IDE+ \cite{zheng2016mars} as the feature learning method. For IDE+, we employ the training strategy in \cite{zhong2018camera}. We adopt ResNet-50 \cite{DBLP:conf/cvpr/HeZRS16} pre-trained on ImageNet \cite{imagenet} as the backbone network. All the images are resized to $256 \times 128$. 
During training, we adopt random flipping and random cropping as data augmentation methods. Dropout probability is set to 0.5. The initial learning rate is set to 0.001 for the layers in the backbone network, and to 0.01 for the remaining layers. The learning rate is decayed by 10 after 40 epochs. \textcolor{black}{We use mini-batch SGD to train IDE+ on a Tesla K80 GPU in a total of 60 epochs. Training parameters such as batch size, momentum, and gamma are set to 16, 0.9, and 0.1, respectively.} We do not fine-tune the batch normalization \cite{IoffeS15} layers. 
During testing, given an input image, we extract the 2,048-dim Pool5 vector for retrieval under the Euclidean distance.

\textbf{SPGAN training and testing.} SPGAN consists of CycleGAN and SiaNet. For CycleGAN, we adopt the architecture released by its authors \cite{cycle}. We use instance normalization \cite{innorm} for generators but no normalization for the discriminators. For SiaNet, it contains 3 convolutional layers, 3 max pooling layers and 2 fully connected (FC) layers, configured as below. (1) Conv. $4 \times 4$, stride = 2, \#feature maps = 64; (2) Max pooling $2 \times 2$, stride = 2; (3) Conv. $4 \times 4$, stride = 2, \#feature maps = 128; (4) Max pooling $2 \times 2$, stride = 2;  (5) Conv. $4 \times 4$, stride = 2, feature maps = 256; (6) Max pool $2 \times 2$, stride = 2;  (7) Max pooling $2 \times 2$, stride = 2; (8) FC, output dimension = 256; 9) FC, output dimension = 128.

 SPGAN is an unsupervised method, \ie, we do not use any ID annotation during the training. In all experiment, we empirically set $ \beta=5, \gamma=2$ in Eq. \ref{spgan_Objective},  $m=2$ in Eq. \ref{eq:Contrastive}, and  $\alpha=10$ in Eq. \ref{cycleloss}. The input images are resized to $256 \times 128$. During training, two data augmentation methods, random flipping and random cropping, are employed. We use the Adam optimizer \cite{Adam} with a batch size of 1, and the $\beta_{1}$ and $\beta_{2}$ are set to 0.5 and 0.999, respectively. The  initial learning rate is 0.0002, and we stop training after 6 epochs. During testing, we employ the Generator $G$ for the source (Market) $\to$ target (Duke) image translation and the Generator $F$ for the target (Duke) $\to$ source (Market) image translation. 

With translated images, we use three strategies to learn a re-ID model: (1) using translated images as training data; (2) using original images and translated images as training data; (3) using translated images to fine-tune the model trained on source images. The results of the three methods are nearly the same, and we adopt the third one to train re-ID models in all the experiment.

\textbf{eSPGAN training and testing.} eSPGAN consists of two models: an image translator and a feature learner. In this paper, we adopt CycleGAN as the image translator and IDE+ as the feature learner, and follow their original architectures. The input images are all resized to $256 \times 128$. Besides, the feature learner is pre-trained on the source dataset following the above setting of feature learning method. During the training eSPGAN, we use two data augmentations: random flipping and random cropping. We set the batch size to $16$. 
For image translator, we use Adam optimizer. The learning rate is $0.0001$ at the first $10$ epochs and linearly decays to $0$ in the remaining 5 epochs.
For feature learner, we use mini-batch SGD in a total of 15 epochs. 
The initial learning rate is set to 0.001 for the layers in the backbone network, and to 0.01 for the remaining layers. The learning rate is decayed by 10 after 10 epochs.
For all the experiment, we set hyper-parameters following CycleGAN for simplicity. Besides, we set the $\lambda=5$ in Eq. \ref{eSPGAN_Objective}. Note that the image translator (CycleGAN) and the feature learner (IDE+) are trained end-to-end, so they share the same image preprocessing procedure. Specifically, we normalize the image with the same mean (0.5, 0.5, 0.5) and standard deviation (0.5, 0.5, 0.5) for both the image translator and the feature learner. At the test time, the re-ID model produced by the feature learner is directly used for the target dataset.

\setlength{\tabcolsep}{8pt}
\begin{table*}[!htp]
\caption{Comparison of various methods on the target domains. When tested on Duke, Market is used as the source dataset, and vice versa. ``Supervised Learning'' denotes using labeled training images on the corresponding target dataset. ``Direct Transfer'' means directly applying the source-trained model on the target domain.  \textcolor{black}{``Direct Transfer (ColorJitter)'' means using images of randomly increased/decreased brightness, contrast, and saturation during training.} When local max pooling (LMP) is applied, the number of parts is set to 8. We use IDE + \cite{zheng2016mars} for feature learning.}
\begin{center}
\begin{tabular}{|l|ccccc|ccccc|}
\hline
\multicolumn{1}{|c|}{\multirow{2}{*}{Methods}}&\multicolumn{5}{c|}{DukeMTMC-reID}&\multicolumn{5}{c|}{Market-1501}\\
\cline{2-11}
\multicolumn{1}{|c|}{}&rank-1&rank-5&rank-10&rank-20&mAP&rank-1&rank-5&rank-10&rank-20&mAP\\
\hline
\hline
Supervised Learning &76.5&87.5&91.1 & 93.6&58.9& 85.1 &94.4&96.6&97.8&66.3\\
\hline
Direct Transfer &38.4&54.3&61.0 &66.1&22.0 &48.1&66.3&73.1&79.0&21.2\\
\textcolor{black}{Direct Transfer (ColorJitter)} & \textcolor{black}{41.0}& \textcolor{black}{56.4}& \textcolor{black}{63.0} &\textcolor{black}{67.5}& \textcolor{black}{23.0}& \textcolor{black}{51.0}&\textcolor{black}{67.6} & \textcolor{black}{73.5}&\textcolor{black}{80.5}&\textcolor{black}{22.1} \\
\hline
CycleGAN (basel.) &40.2&56.7&62.8&68.2&22.4&51.6&68.1&75.8&81.5&22.3\\
CycleGAN (basel.) + $L_{ide}$ &42.5&58.5&64.3&69.3&23.1& 53.0&70.2&77.6&82.4&23.5 \\
\hline
SPGAN ($m=2$)&{44.3}&{61.2}&{66.0}&{71.1}&{24.6}&{54.6}&{72.4}&{79.7}&{84.2}&{25.1}\\
{SPGAN ($m=2$) + LMP}&47.1&{63.8}&{70.0}&{74.2 }&{26.1}&{57.2}&{74.0}&{82.1}&{86.4 }&{27.4} \\
\hline
eSPGAN&47.9&61.9&67.1&73.2&26.1&59.5&76.0&82.2&88.2&28.9 \\
{eSPGAN+ LMP}&\textbf{52.6}&\textbf{66.3}&\textbf{71.7}&\textbf{76.2}&\textbf{30.4}&\textbf{63.6}&\textbf{80.1}&\textbf{86.1}&\textbf{90.1}&\textbf{31.7} \\
\hline
\end{tabular}
\end{center}
\setlength{\abovecaptionskip}{0cm} 
\setlength{\belowcaptionskip}{0cm} 
 \label{table:baselines}
\end{table*}

\subsection{Baseline Evaluation}
In this section, we evaluate the direct transfer method and the ``learning via translation'' baseline. 

\textbf{Dataset bias in re-ID.} To demonstrate the influence of the dataset bias, we report the results of the supervised learning method and the direct transfer method in Table \ref{table:baselines}. The supervised learning method  is trained and tested on the same domain, which defines the upper bound of our system. In the direct transfer, we train a re-ID model on the source domain and directly deploy the resulting model on the target domain without any domain adaptation technique. We clearly observe a large performance drop when directly using a source-trained model on the target domain. For example, the IDE+ model trained and tested on Market achieves a rank-1 accuracy of $85.1\%$, but drops to $48.1\%$ when trained on Duke and tested on Market. A similar drop can be observed when Duke is used as the target domain, which is consistent with the experiment reported in \cite{fan17unsupervised}. The reason behind the performance drop is the large difference between data distributions in different domains.

\textbf{Effectiveness of the ``learning via translation'' baseline.} We use CycleGAN as the baseline for source-target image translation. It is worth noting that CycleGAN does not involve any identity-preserving technique. As shown in Table \ref{table:baselines}, the baseline effectively improves over the direct transfer method on the target dataset. For example, comparing with the direct transfer, the CycleGAN baseline gains $+3.5\%$ improvement in rank-1 accuracy on Market. 

Moreover, when we adopt the inside-domain identity loss in CycleGAN, we observe some further improvement. When tested on Duke and Market, the rank-1 accuracy gains brought by adding the identity loss are +2.3\% and +1.4\%, respectively. We speculate that the inside-domain identity loss constrains the mapping functions, such that some original semantics are preserved in the translated images. To some extent, the effectiveness of the inside-domain identity loss suggests the necessity of preserving image content. Overall, considering the results of the baselines using CycleGAN and CycleGAN + inside-domain identity loss, we conclude that the ``learning via translation'' baseline is effective in domain adaptation. However, comparing with SPGAN and eSPGAN, its effectiveness is limited without learning the identity-preserving property. 

\textcolor{black}{\textbf{Style change.} Our methods perform distribution alignment in raw pixel space - translating source data to the ``style” of a target domain. The ``style” change is complex and abstract; it involves many various factors. For example, from the visual examples in Fig. \ref{samples} and Fig. \ref{fig:sample_images}, we observe that resolution, illumination, color, and background are changed, as well as other changes that are harder to describe.}

\begin{figure}[t]
\setlength{\abovecaptionskip}{0cm} 
\setlength{\belowcaptionskip}{0cm}
\begin{center}
\includegraphics[width=1.0\linewidth]{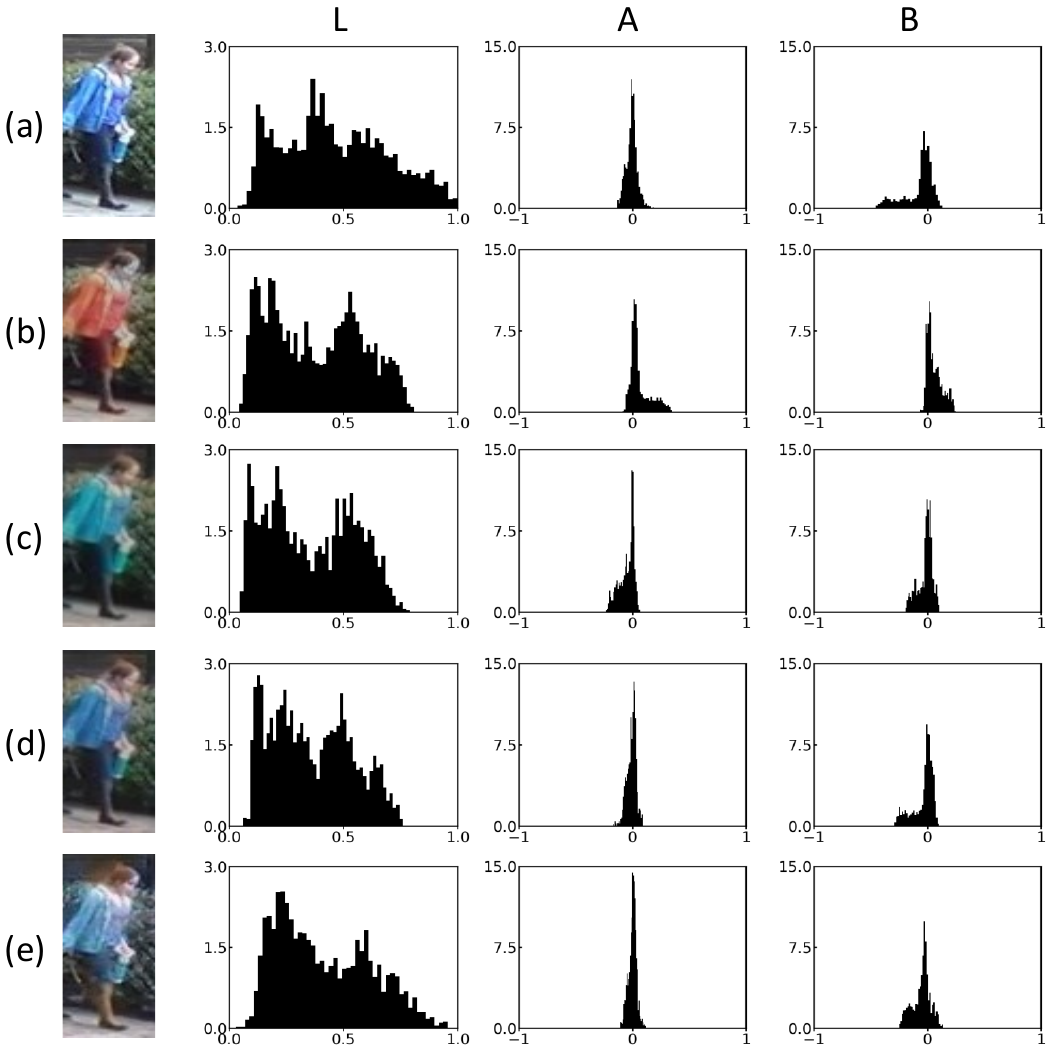}
\end{center}
\caption{\textcolor{black}{Global histograms of images in LAB color space. From top to bottom: (a) original image, outputs of (b) CycleGAN, (c) CycleGAN + Lide, (d) SPGAN and (e) eSPGAN, respectively. The LAB space expresses color with three values: $L^{*}$ for the luminance from black (0) to white (1), $A^{*}$ from green (-1) to red (+1), and $B^{*}$ from blue (-1) to yellow (+1). The histogram shifts in three channels demonstrate that illumination and color composition are the changed factors introduced by image translation.}}
\label{fig:lab}
\end{figure}

\textcolor{black}{We provide a quantitative analysis of changes in two example visual factors: illumination and color. In Fig. \ref{fig:lab}, we visualize channel-wise histograms of translation examples in the LAB color space. We choose the LAB color space because it relates closely to how human vision works \cite{palacio2018deep}.}
\textcolor{black}{We observe all methods introduce a distribution shift of the ``L” channel (the distribution moves towards left), making more areas of image dark. Moreover, all the compared methods introduce the distribution shifts in ``A” and ``B” channels, too. These shifts correspond to the color composition changes. For example, CycleGAN introduces the largest distribution shifts in channels ``A” and ``B”; it changes the color from blue to red. The histogram shifts in the LAB space demonstrate that both illumination and color composition are the changed factors introduced by image translation.}

\textcolor{black}{To further study the impact of illumination, we have newly added a data augmentation method (``colorjitter”) to the direct transfer baseline. ``Colorjitter” uses images of randomly increased/decreased brightness, contrast, and saturation. As shown in Table \ref{table:baselines}, ``colorjitter” augmentation brings about some improvement over the direct transfer baseline. This indicates illumination is a factor that causes the dataset bias between Duke and Market. However, even with ``colorjitter”, the direct transfer baseline is still lower than CycleGAN + $L_{ide}$, SPGAN and eSPGAN. It means that considering only the illumination during image translation is insufficient. More importantly, it suggests that manually designing how certain factors should be changed is not optimal. In fact, SPGAN and eSPGAN not only consider multiple factors (\eg, color composition, background, and some indescribable ones), but also change them in an automatic manner, which is much more effective than manually designed changes.}

\textcolor{black}{In addition, the ``style” is an abstract and comprehensive notion, and it is non-trivial to list and define all the related factors. Thus, we cannot manually specify certain factor changes for the image translator to learn. In comparison, our GAN-based method looks at the global data distribution, such that image “style” is optimized/changed automatically.}

\begin{figure}[!t]
\setlength{\abovecaptionskip}{0cm} 
\setlength{\belowcaptionskip}{0cm}
\begin{center}
\includegraphics[width=1\linewidth]{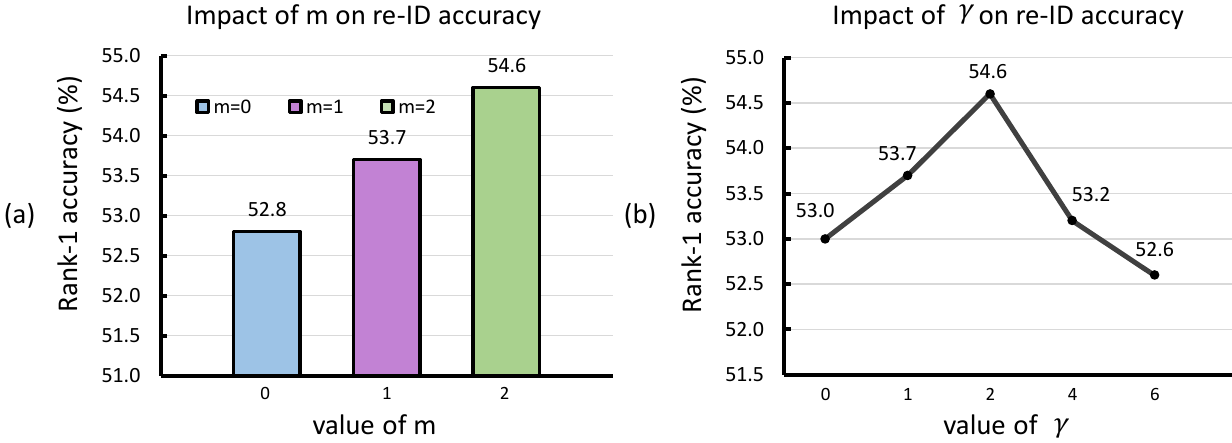}
\end{center}
\caption{{The impact of the hyper-parameters of SPGAN on the re-ID rank1 accuracy. (a): the impact of $m$ in Eq. \ref{eq:Contrastive}, a larger $m$ means that the loss of negative training samples has a higher weight in back-propagation. (b): the impact of $\gamma$ in Eq. \ref{spgan_Objective}, a larger $\gamma$ means a larger weight of similarity preserving constraint. The results are on Market.}}
\label{spgan_para}
\end{figure}

\subsection{Evaluation of SPGAN}
On top of the ``learning via translation" baseline, we replace CycleGAN with SPGAN and leave the feature learning component unchanged. In this section, we present a step-by-step evaluation and analysis of SPGAN.

\textbf{SPGAN effect.} On top of the ``learning via translation" baseline, we replace CycleGAN with SPGAN ($m=2$). The effectiveness of the proposed similarity preserving constraint can be seen in Table \ref{table:baselines}. On Duke, the similarity preserving constraint leads to $+1.8\%$ and $+1.5\%$ improvements over CycleGAN + $L_{ide}$ in rank-1 accuracy and mAP, respectively. On Market, the performance gains are $+1.6\%$ and $1.6\%$. The working mechanism of SPGAN consists in preserving the underlying visual cues associated with the ID labels. The consistent improvement suggests that this working mechanism is critical for generating suitable samples for training re-ID models in the target domain. Examples of translated images by SPGAN are shown in Fig. \ref{fig:sample_images}.

\textbf{Sensitivity of SPGAN to key hyper-parameters.} SPGAN has two parameters that affect the re-ID accuracy, \ie,  $m$ in Eq. \ref{eq:Contrastive} and $\gamma$ in Eq. \ref{spgan_Objective}. We conduct the experiment to analyze the impact of the $m$ and $ \gamma$ on Market, and results are shown in Fig. \ref{spgan_para}.

First, $m \in [0, 2]$ is the margin that defines the separability of negative pairs in the embedding space. If $m=0$, the loss of the negative pairs is not back-propagated. If $m$ gets larger, the weight of negative pairs in loss calculation increases. When turning off the contribution of negative pairs ($m=0$), SPGAN only marginally improves the accuracy on Market. When increasing $m$ to 2, we have much superior accuracy. It indicates that the negative pairs are critical to the system. 

Second, $\gamma$ controls the relative importance of the proposed similarity preserving constraint. As shown in Fig. \ref{spgan_para} (b), the proposed constraint is proven effective when compared to $\gamma=0$, but a larger $\gamma$ does not bring more gains in re-ID accuracy. Specifically, $\gamma=2$ yields the best accuracy.
\begin{figure}[t]
\setlength{\abovecaptionskip}{-0.2cm} 
\setlength{\belowcaptionskip}{-0.3cm}
\begin{center}
\includegraphics[width=0.9 \linewidth]{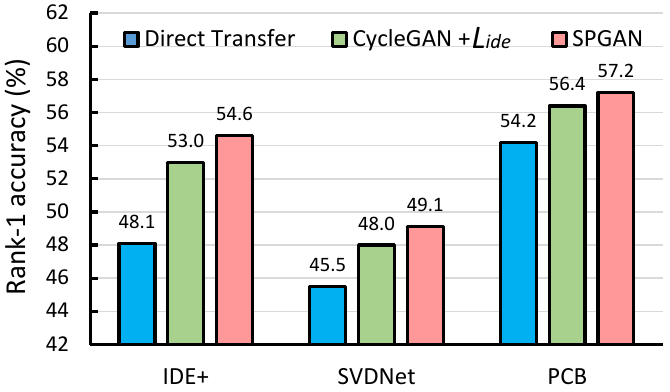}
\end{center}
\caption{Domain adaptation performance with different feature learning methods, including IDE$^+$ \cite{zheng2016mars}, SVDNet \cite{SVD}, PCB \cite{PCB}. Three domain adaptation methods are compared, \ie, direct transfer, CycleGAN with identity loss, and the proposed SPGAN. The results are on Market.}
\label{fig:spgan_methods}
\end{figure}

\textbf{Comparison of different feature learning methods.} Given the same translated images, we evaluate three feature learning methods, \ie, IDE$^+$ \cite{zheng2016mars}, SVDNet \cite{SVD}, PCB \cite{PCB}. We choose Market as the target dataset and duke as the source dataset, and results are shown in Fig. \ref{fig:spgan_methods}. 
Under the domain adaptation setting, we observe that better feature learning methods lead to higher direct transfer results. 
For example, PCB achieves higher accuracy than IDE+ under the supervised setting on Market (92.3\% vs. 85.1\%), and its direct transfer accuracy is also higher than that of IDE+ (54.2\% vs. 48.1\%).
As shown in Fig. \ref{fig:spgan_methods}, the SPGAN gains consistent improvement with three different feature learning methods. Compared with the very high direct transfer accuracy (54.2\%) of PCB, the ``learning via translation" framework baseline (CycleGAN + \text{$L_{ide}$}) gains +2.2\% improvement, and the SPGAN gains $+3.0\%$ improvement.

\subsection{Evaluation of eSPGAN}
\textbf{eSPGAN effect.} An evaluation of eSPGAN is shown in Table \ref{table:baselines}. 
eSPGAN adopts CycleGAN + $L_{ide}$ as the image translator. Compared with CycleGAN + $L_{ide}$, eSPGAN further gains +6.5 \% in rank-1 accuracy on the Market dataset. We also observe the significant improvement on Duke dataset, the rank-1 accuracy increases from 42.5\% to 47.9\%. Moreover, eSPGAN greatly improves the performance of direct transfer, the rank-1 accuracy on Market and Duke increases from 48.1\% and 38.4\% to 59.5\% and 47.9\%, respectively. The experimental results strongly indicate that eSPGAN can effectively leverage the knowledge of image translation and  feature learner to learn more discriminative embeddings for the target domain. 
Examples of translated images by eSPGAN are shown in Fig. \ref{samples}.
\setlength{\tabcolsep}{7pt}
\begin{table}[t]
\caption{Comparision of eSPGAN and Na\"ive eSPGAN on Market and Duke datasets. Rank-1 accuracy (\%) and mAP (\%) are shown.}
\setlength{\belowcaptionskip}{0cm}
\begin{center}
\begin{tabular}{|l|cc|cc|}
\hline
\multicolumn{1}{|l|}{\multirow{2}{*}{Methods}}&\multicolumn{2}{c|}{DukeMTMC-reID}&\multicolumn{2}{c|}{Market-1501}\\
\cline{2-5}
\multicolumn{1}{|c|}{}&Rank-1&mAP&Rank-1&mAP\\
\hline
\hline
CycleGAN + $L_{ide}$ &42.5&23.1&53.0&23.5\\
Na\"ive eSPGAN&44.3&24.4&55.1&24.9\\
eSPGAN &{47.9}&{26.1}&{59.5} &{28.9} \\
\hline
\end{tabular}
\end{center}
\setlength{\abovecaptionskip}{-0.1cm} 
\label{table:naive_espgan}
\end{table}

\begin{figure}[t]
\setlength{\abovecaptionskip}{0cm} 
\setlength{\belowcaptionskip}{0cm}
\begin{center}
\includegraphics[width=0.80\linewidth]{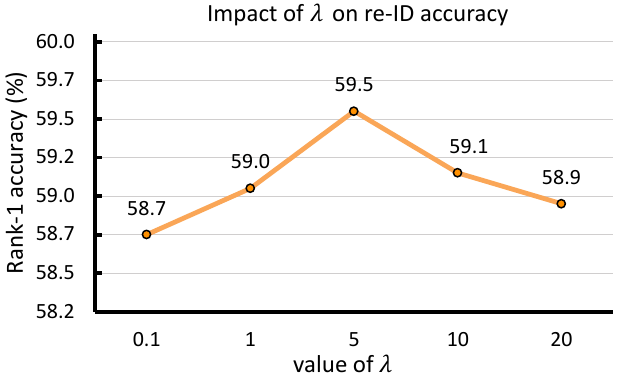}
\end{center}
\caption{Sensitivity of eSPGAN to key parameter $\lambda$ in Eq. \ref{eSPGAN_Objective}. A larger $\lambda$ means that the feature learner has a greater influence on the image translator. The results are on Market.}
\label{fig:espgan_para}
\end{figure}

\textbf{ Na\" ive eSPGAN.} By this we mean that the parameters of feature learner will not be updated during training. Thus, the image translator na\" ively utilizes a pre-trained source model to guide its translation procedure. 
Note that Na\" ive eSPGAN only aims to learn an image translator.
As analyzed in Section \ref{know}, many existing methods adopt this way to preserve the similarity of the generated image \cite{hoffman2018cycada, liu2018pose, chen2018zero, ShiriYPHK19}. In Table \ref{table:naive_espgan}, we compare eSPGAN with Na\" ive eSPGAN. We can observe that Na\" ive eSPGAN can improve the accuracy of the baseline (CycleGAN + $L_{ide}$). However,  the accuracy of Na\" ive eSPGAN is still much lower than eSPGAN. For example, eSPGAN obtains a much higher rank-1 accuracy than Na\" ive eSPGAN (47.9\% vs. 44.3\%) on Duke. This suggests that the parameters of pre-trained feature learner should be updated during training, so that the knowledge of feature learner and image translator can be gradually transferred to each other.

\textbf{ Analysis of the hyper-parameter of eSPGAN.} $\lambda$ in Eq. \ref{eSPGAN_Objective} is an important parameter of eSPGAN, which defines the influence of the feature learner on the image translator. To further analyze the effect of $\lambda$, we vary it from 0.1 to 20 to evaluate the performance of eSPGAN on Market. The rank-1 accuracies when using different $\lambda$ are plotted in  Fig. \ref{fig:espgan_para}. In our system, when the $\lambda$ is set to 5, we can obtain the best re-ID accuracy. Note that setting the $\lambda$ to 0 means the feature learner has no influence on the image translator.

\textbf{ Analysis of the different forms of the feature learner.} eSPGAN consists of an image translator and a feature learner. The feature learner is crucial for the image translator to generate similarity-preserving images, \ie, the translated image maintains its visual contents that associated with the identity information. We analyze two forms of the feature learner, \ie, IDE$^+$ \cite{zheng2016mars}, PCB \cite{PCB}. We choose Market as the target dataset and duke as the source dataset and report results in Fig. \ref{fig:espgan_methods}. Under the domain adaptation setting, we observe that eSPGAN gains consistent improvement with two feature learning methods. For example, when using PCB as feature learning method, eSPGAN gains +4.8\% improvement over the CycleGAN + $L_{ide}$.

\setlength{\tabcolsep}{5pt}
\begin{table}[!t]
\caption{Performance of eSPGAN after source-target adaptation on the \textbf{source} dataset.  Rank-1 accuracy (\%) and mAP (\%) are shown.}
\setlength{\belowcaptionskip}{-1cm}
\begin{center}
\begin{tabular}{|l|cc|cc|}
\hline
\multicolumn{1}{|l|}{\multirow{2}{*}{Methods}}&\multicolumn{2}{c|}{DukeMTMC-reID}&\multicolumn{2}{c|}{Market-1501}\\
\cline{2-5}
\multicolumn{1}{|c|}{}&Rank-1&mAP&Rank-1&mAP\\
\hline
\hline
Supervised Learning &76.5&58.9&85.1&66.3\\
eSPGAN &{76.1}&{57.7}&{84.6} &{65.4} \\
\hline
\end{tabular}
\end{center}
\setlength{\abovecaptionskip}{-0.1cm} 
\label{table:real sample}
\end{table}

\begin{figure}[t]
\setlength{\abovecaptionskip}{0cm} 
\setlength{\belowcaptionskip}{0cm}
\begin{center}
\includegraphics[width=0.85\linewidth]{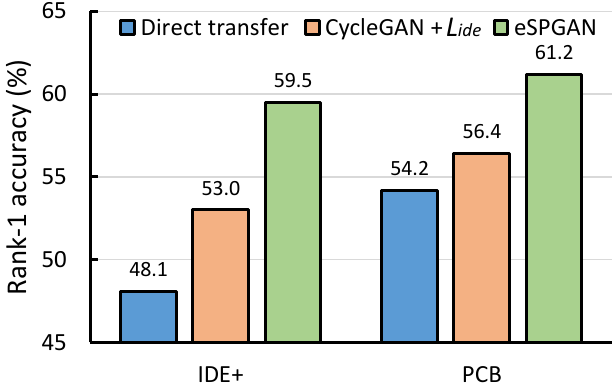}
\end{center}
\caption{eSPGAN performance with different feature learning methods, including IDE$^+$ \cite{zheng2016mars}, PCB \cite{PCB}. Three domain adaptation methods are compared, \ie, direct transfer, CycleGAN with identity loss, and eSPGAN. The results are on Market.}
\label{fig:espgan_methods}
\end{figure}
\begin{figure}[t]
\setlength{\abovecaptionskip}{-0.1cm} 
\setlength{\belowcaptionskip}{-0.1cm}
\begin{center}
\includegraphics[width=1 \linewidth]{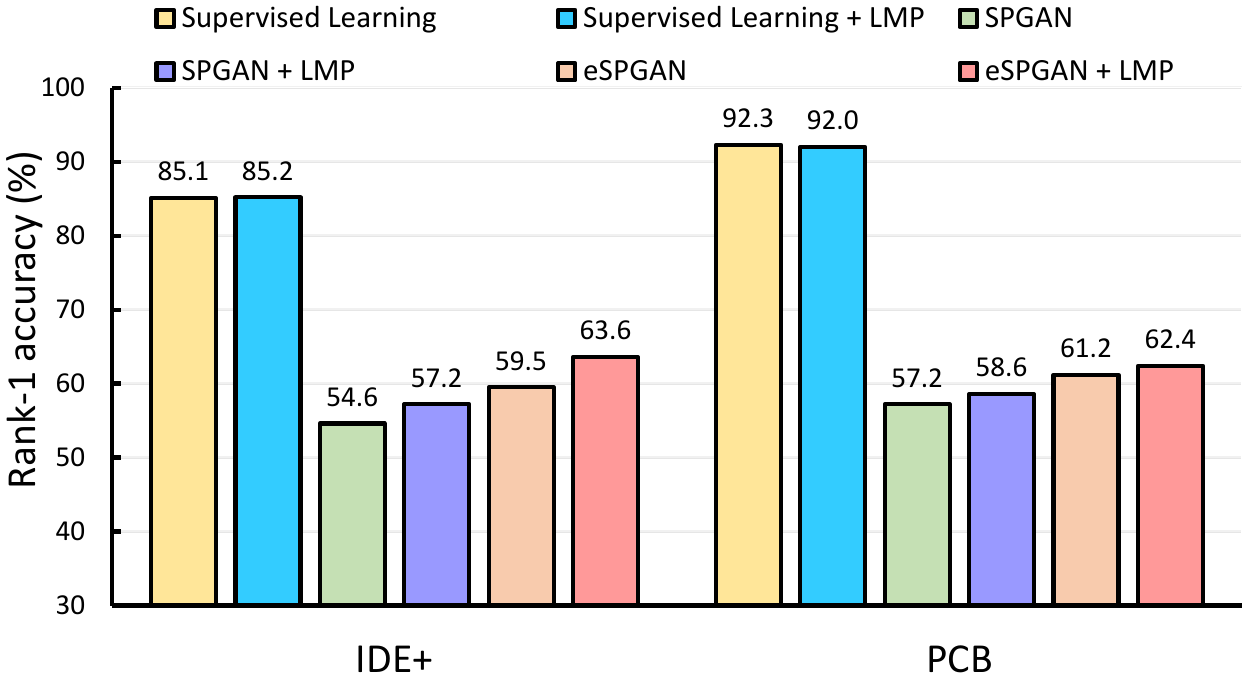}
\end{center}
\caption{The experiment of LMP $(P=8)$ on scenarios of supervised learning and domain adaptation with SPGAN and eSPGAN. Two feature learning methods are compared, \ie, IDE$^+$ \cite{zheng2016mars}, and PCB \cite{PCB}. The results are on Market.}
\label{fig:lmp_method}
\end{figure}

\textbf{Domain invariant person embeddings.} As discussed in Section \ref{know}, we also use source images when training eSPGAN. This practice ensures feature learner will not be led to divergence by the poorly translated images. 
In addition, using both source and translated images for feature learning leads to domain invariant person embeddings. 
To validate this, We also report the accuracy of eSPGAN after source-target adaptation on the \textbf{source} dataset in Table \ref{table:real sample}.
We observe eSPGAN slightly decreases the rank-1 accuracy on the source dataset after source-target adaptation. Moreover, compared with the direct transfer baseline, eSPGAN significantly improves the performance on the target dataset. Thus, eSPGAN can learn person embeddings that are effective for both the source and target datasets.

\textbf{Impact of two heuristic constraints.} We further investigate the impact of two heuristic constraints of SPGAN on eSPGAN. We add the two heuristic constraints to eSPGAN during training, and report results in Table \ref{table:constraits}. We can observe that two heuristic constraints do not improve the performance of eSPGAN. This is because the feature learner provides the sufficiently informative and accurate constraint for the image translator. Thus, eSPGAN does not adopt two heuristic constraints during training.

\textbf{Local max pooling.} We apply the LMP on the last convolution layer of a re-ID model to mitigate the influence of noise. Note that LMP is directly adopted in the feature extraction step for testing without any fine-tuning. In Table \ref{table:baselines}, we can observe that LMP (P=8) can improve the accuracy of SPGAN and eSPGAN. With the help of LMP (P=8), SPGAN obtains +2.6\% improvement on Market in rank-1 accuracy.  LMP also improves the rank-1 accuracy of eSPGAN from 59.5\% to 63.6\% on Market.
We empirically study how the number of parts and the pooling mode affect the accuracy. The experiment is conducted on eSPGAN. The performance of various numbers of parts ($P = 1, 2, 4, 8$) and different pooling modes (max or average) is provided in Table \ref{table:LMP}. When using average pooling and $P=1$, we have the original GAP used in ResNet-50. From these results, we speculate that with more parts, a finer partition leads to higher discriminative descriptors and thus higher re-ID accuracy. 

Moreover, we test LMP on supervised learning and domain adaptation scenarios with two feature learning methods, \ie, IDE$^+$ \cite{zheng2016mars} and PCB \cite{PCB}. As shown in Fig. \ref{fig:lmp_method}, LMP does not guarantee stable improvement on supervised learning as observed in ``IDE$^+$'' and PCB.
However, when applied in the scenario of domain adaptation, LMP yields consistent improvement over IDE$^+$ and PCB.
We believe that the superiority of LMP probably lies in that it could filter out some detrimental signals in the descriptor induced by unsatisfied translated images.

\setlength{\tabcolsep}{3pt}
\begin{table}[t]
\caption{Impact of two heuristic constraints on eSPGAN.  Rank-1 accuracy (\%) and mAP (\%) are shown.}
\setlength{\belowcaptionskip}{-1cm}
\begin{center}
\begin{tabular}{l|c|cc|cc}
\hline
\multicolumn{1}{l|}{\multirow{2}{*}{ }}&{\multirow{2}{*}{\shortstack{Training w/ \\ heuristic constraints?}}}&\multicolumn{2}{c}{DukeMTMC-reID}&\multicolumn{2}{c}{Market-1501}\\
\cline{3-6}
\multicolumn{1}{c|}{}&&Rank-1&mAP&Rank-1&mAP\\
\hline
\hline
eSPGAN &&{47.9}&{26.1}&{59.5} &{28.9} \\
eSPGAN &\checkmark&47.5&26.2&59.6&28.6\\
\hline
\end{tabular}
\end{center}
\setlength{\abovecaptionskip}{-0.1cm} 
\label{table:constraits}
\end{table}
\setlength{\tabcolsep}{6pt}
\begin{table}[t]
\caption{Performance of various pooling strategies with different numbers  of parts ($P$) and pooling modes (maximum or average) over eSPGAN. 
}
\setlength{\belowcaptionskip}{-0.7cm}
\begin{center}
\begin{tabular}{|l|c|c|cc|cc|}
\hline
\multicolumn{1}{|l|}{\multirow{2}{*}{\#parts}}&{\multirow{2}{*}{mode}}&{\multirow{2}{*}{dim}}&\multicolumn{2}{c|}{DukeMTMC-reID}&\multicolumn{2}{c|}{Market-1501}\\
\cline{4-7}
\multicolumn{1}{|c|}{}& & &rank-1&mAP&rank-1&mAP\\
\hline 
\hline
\quad\multirow{2}{*}{1}&Avg&\multirow{2}{*}{2048}&47.9 & 26.1& 59.5&28.9 \\
&Max& &50.7&28.1&62.6&30.2\\
\hline 
\quad{\multirow{2}{*}{2}}&Avg&\multirow{2}{*}{4096}& 50.1&27.6& 61.3 &29.8  \\
&Max& &51.9&28.5&62.9& 30.5\\
\hline 
\quad{\multirow{2}{*}{4}}&Avg&\multirow{2}{*}{8192}& 50.1& 28.0&62.5&30.1 \\
&Max& &52.4&29.0&63.2& 30.9\\
\hline 
\quad{\multirow{2}{*}{8}}&Avg&\multirow{2}{*}{16384}&51.5&28.9&63.2&31.0\\
&Max& &\textbf{52.6} &\textbf{29.6} & \textbf{63.6} &\textbf{31.7}   \\
\hline
\end{tabular}
\label{table:LMP}
\end{center}
\end{table}
\setlength{\tabcolsep}{6pt}
\begin{table}[t]
\caption{Comparison with the state-of-the-art methods on Market. ``SQ'' and ``MQ'' are the single-query and multiple-query settings, respectively. 
}
\begin{center}
\begin{tabular}{|l|c|cccc|}
\hline
\multicolumn{1}{|l|}{\multirow{2}{*}{Methods}}&\multicolumn{5}{c|}{Market-1501}\\
\cline{2-6}
\multicolumn{1}{|c|}{}&Setting&Rank-1&Rank-5&Rank-10&mAP\\
\hline
\hline
Bow \cite{DBLP:conf/iccv/ZhengSTWWT15}&SQ&35.8&52.4&60.3&14.8\\
LOMO \cite{DBLP:conf/cvpr/LiaoHZL15}&SQ&27.2&41.6&49.1&8.0\\
\hline
UMDL \cite{DBLP:conf/cvpr/PengXWPGHT16}&SQ&34.5&52.6&59.6&12.4\\
PUL \cite{fan17unsupervised}&SQ&45.5&60.7&66.7&20.5\\
Direct transfer &SQ&48.1&66.3&73.1&21.2\\
Direct transfer &MQ&52.3&70.1&77.2&25.0\\
CAMEL \cite{CAMEL} &MQ&54.5&-&-&26.3\\
TJ-AIDL \cite{wang2018}&SQ&58.2&74.8&81.1&26.5\\
PTGAN \cite{ptgan}&SQ&38.6&-&66.1&-\\
HHL \cite{zhong2018generalizing}&SQ&62.2&78.8&84.0&31.4\\
\hline
SPGAN&SQ&54.6&71.4&79.1&25.1\\
SPGAN&MQ&58.0&74.7&83.2&29.6\\
{SPGAN+LMP}&SQ&57.2&74.0&82.1&27.4\\
\hline
eSPGAN&SQ&59.5&76.0&82.2&28.9\\
eSPGAN&MQ&\textbf{63.5}&\textbf{81.1}&\textbf{87.3}&\textbf{34.5}\\
{eSPGAN+LMP}&SQ&\textbf{63.6}&\textbf{80.1}&\textbf{86.1}&\textbf{31.7}\\
\hline
\end{tabular}
\end{center}
\setlength{\abovecaptionskip}{-0cm} 
\label{table:sota-market}
\end{table}

\subsection{Comparison with State-of-the-art Methods}
Finally, we compare SPGAN and eSPGAN with the state-of-the-art unsupervised learning methods on Market and Duke in Table \ref{table:sota-market} and Table \ref{table:sota-duke}, respectively.

\setlength{\tabcolsep}{10pt}
\begin{table}[t]
\caption{Comparison with the state-of-the-art methods on Duke under the single-query setting. 
}
\begin{center}
\begin{tabular}{|l|cccc|}
\hline
\multicolumn{1}{|l|}{\multirow{2}{*}{Methods}}&\multicolumn{4}{c}{DukeMTMC-reID}\\
\cline{2-5}
\multicolumn{1}{|c|}{}&Rank-1&Rank-5&Rank-10&mAP\\
\hline
\hline
Bow \cite{DBLP:conf/iccv/ZhengSTWWT15}&17.1&28.8&34.9&8.3\\
LOMO \cite{DBLP:conf/cvpr/LiaoHZL15}&12.3&21.3&26.6&4.8\\
\hline
UMDL \cite{DBLP:conf/cvpr/PengXWPGHT16}&18.5&31.4&37.6&7.3\\
Direct transfer &38.4&54.3&61.0&22.0\\
PUL \cite{fan17unsupervised}&30.0&43.4&48.5&16.4\\
PTGAN \cite{ptgan}&27.4&-&50.7&-\\
TJ-AIDL \cite{wang2018}&44.3&59.6&65.0&23.0\\
HHL \cite{zhong2018generalizing}&46.9&61.0&66.7&27.2\\
\hline
SPGAN& {44.3}& {61.2}& {66.0}& {24.6} \\
{SPGAN+LMP}&47.1&63.8&70.0&26.1\\
\hline
eSPGAN &{47.9}&{61.9}&{67.1} &{26.1} \\
{eSPGAN+LMP}&\textbf{52.6}&\textbf{66.3}&\textbf{71.7}&\textbf{29.6}\\
\hline
\end{tabular}
\end{center}
\setlength{\abovecaptionskip}{-0.1cm} 
\label{table:sota-duke}
\end{table}
\textbf{Market as the target domain.} On Market, we first compare the proposed methods with two hand-crafted features, \ie, bag-of-Words (BoW) \cite{DBLP:conf/iccv/ZhengSTWWT15} and local maximal occurrence (LOMO) \cite{DBLP:conf/cvpr/LiaoHZL15}. These two hand-crafted features are directly applied to the target dataset without any training process, their inferiority can be clearly observed.
We also compare with existing unsupervised learning methods, including the clustering-based asymmetric metric learning (CAMEL) \cite{CAMEL}, the Progressive Unsupervised Learning (PUL) \cite{fan17unsupervised}, and UMDL \cite{DBLP:conf/cvpr/PengXWPGHT16}. For UMDL, we use the results reproduced by Fan \emph{et al.} \cite{fan17unsupervised}. 
Moreover, we compare the proposed methods with recent domain adaptation methods of re-ID, \ie, PTGAN \cite{ptgan}, TJ-AIDL \cite{wang2018}, and HHL \cite{zhong2018generalizing}.
In the multiple-query setting, SPGAN and eSPGAN arrive at rank-1 accuracy = 58.0\% and 63.5\%, respectively.
The accuracy of SPGAN is 3.5\% higher than CAMEL \cite{CAMEL}.
In the single-query setting, SPGAN achieves 54.6\% in  rank-1 accuracy, and eSPGAN achieves 59.5\%. We can observe that SPGAN outperforms many other methods. With the help of  LMP (P=8), SPGAN is comparable with recent work TJ-AIDL \cite{wang2018}. Moreover, eSPGAN outperforms TJ-AIDL \cite{wang2018} by 1.3\%, which indicates that it is beneficial to jointly optimize feature learner and image translator. 
With the help of  LMP (P=8), eSPGAN achieves a new state-of-the-art rank-1 accuracy=63.6\%, which is 1.4\% higher than the second best method HHL \cite{zhong2018generalizing}.
The comparisons indicate the competitiveness of SPGAN and eSPGAN on Market.

\textbf{Duke as the target domain.} On Duke, we compare the results with BoW \cite{DBLP:conf/iccv/ZhengSTWWT15}, LOMO \cite{DBLP:conf/cvpr/LiaoHZL15}, UMDL \cite{DBLP:conf/cvpr/PengXWPGHT16}, and PUL \cite{fan17unsupervised} under the single-query setting (there is no multiple-query setting in DukeMTMC-reID). We also compare with recent domain adaptation methods of re-ID, \ie, PTGAN \cite{ptgan}, TJ-AIDL \cite{wang2018}, and HHL \cite{zhong2018generalizing}.
The result obtained by SPGAN is rank-1 accuracy = 44.3\%, mAP = 24.6\%, which is competitive with the recent work TJ-AIDL \cite{wang2018}. With the help of LMP (P=8), SPGAN is comparable with HHL \cite{zhong2018generalizing}.
Moreover, eSPGAN gains rank-1 accuracy=47.9\%, which is +1\% higher than HHL \cite{zhong2018generalizing}. With the help of LMP (P=8), eSPGAN achieves a new state-of-the-art rank-1 accuracy=52.6\%. 
Therefore, the superiority of SPGAN and eSPGAN can be concluded. 

\section{Conclusion}\label{sec:Conclusion and future work}
This paper focuses on domain adaptation in person re-ID. When models trained on one dataset are directly transferred to another dataset, the re-ID accuracy drops dramatically due to dataset bias. To achieve improved performance in the new dataset, we present a ``learning via translation'' framework characterized by 1) unsupervised image-image translation and 2) supervised feature learning. 
We propose that the underlying (latent) ID information for the foreground pedestrian should be preserved after image-image translation.
 \textcolor{black}{To meet this requirement tailored for re-ID, we propose a similarity preserving generative adversarial network (SPGAN) and its upgraded version, eSPGAN.}
Both aiming at similarity preserving, SPGAN enforces this property by heuristic constraints, while eSPGAN does so by leveraging the discriminative knowledge of the re-ID model. 
We show that SPGAN and eSPGAN better qualify the generated images for domain adaptation and achieve state-of-the-art results on two large-scale person re-ID datasets. 
\textcolor{black}{In the future, we plan to further study the relation between generative and discriminative learning, and improve our method for more general applications in visual understanding.}

\bibliographystyle{IEEEtran}
\bibliography{IEEEabrv,REFS}
\end{document}